\documentclass[sn-mathphys,Numbered]{sn-jnl}

\usepackage{graphicx}%
\usepackage{multirow}%
\usepackage{amsmath,amssymb,amsfonts}%
\usepackage{amsthm}%
\usepackage{mathrsfs}%
\usepackage[title]{appendix}%
\usepackage{xcolor}%
\usepackage{textcomp}%
\usepackage{manyfoot}%
\usepackage{booktabs}%
\usepackage{algorithm}%
\usepackage{algorithmicx}%
\usepackage{algpseudocode}%
\usepackage{listings}%

\usepackage{xcolor, soul}
\sethlcolor{yellow}

\begin{document}

\title[Preop Model Update]{An Endoscopic Chisel: Intraoperative Imaging Carves 3D Anatomical Models}

\author*[1]{\fnm{Jan Emily} \sur{Mangulabnan}} \email{jmangul1@jh.edu}

\author[1]{\fnm{Roger D.} \sur{Soberanis-Mukul}}

\author[1]{\fnm{Timo} \sur{Teufel}}

\author[1]{\fnm{Manish} \sur{Sahu}}
\author[2]{\fnm{Jose L.} \sur{Porras}}
\author[1]{\fnm{S. Swaroop} \sur{Vedula}}
\author[2]{\fnm{Masaru} \sur{Ishii}}
\author[1]{\fnm{Gregory} \sur{Hager}}
\author[1,2]{\fnm{Russell H.} \sur{Taylor}}
\author[1,2]{\fnm{Mathias} \sur{Unberath}}

\affil*[1]{\orgname{Johns Hopkins University}, \orgaddress{\city{Baltimore}, \postcode{21211}, \state{MD}, \country{USA}}}

\affil[2]{\orgname{Johns Hopkins Medical Institutions}, \orgaddress{\city{Baltimore}, \postcode{21287}, \state{MD}, \country{USA}}}





\abstract{
\textbf{Purpose:}
Preoperative imaging plays a pivotal role in sinus surgery where CTs offer patient-specific insights of complex anatomy, enabling real-time intraoperative navigation to complement endoscopy imaging. However, surgery elicits anatomical changes not represented in the preoperative model, generating an inaccurate basis for navigation during surgery progression.
\textbf{\\*Methods:}
We propose a first vision-based approach to update the preoperative 3D anatomical model leveraging intraoperative endoscopic video for navigated sinus surgery where relative camera poses are known. We rely on comparisons of intraoperative monocular depth estimates and preoperative depth renders to identify modified regions. The new depths are integrated in these regions through volumetric fusion in a truncated signed distance function representation to generate an intraoperative 3D model that reflects tissue manipulation.
\textbf{\\*Results:}
We quantitatively evaluate our approach by sequentially updating models for a five-step surgical progression in an ex vivo specimen. We compute the error between correspondences from the updated model and ground-truth intraoperative CT in the region of anatomical modification. The resulting models show a decrease in error during surgical progression as opposed to increasing when no update is employed.
\textbf{\\*Conclusion:}
Our findings suggest that preoperative 3D anatomical models can be updated using intraoperative endoscopy video in navigated sinus surgery. Future work will investigate improvements to monocular depth estimation as well as removing the need for external navigation systems. The resulting ability to continuously update the patient model may provide surgeons with a more precise understanding of the current anatomical state and paves the way toward a digital twin paradigm for sinus surgery.
}




\maketitle

\section{Introduction}\label{sec:introduction}

\noindent \textbf{Background:}  
Chronic sinusitis (CRS) is a common condition affecting 5-12\% of the population where patients experience persistent nasal obstruction, congestion, and drainage ultimately affecting their quality of life~\cite{clark2018nasal,moubayed2022evaluation,lee2021association, wyler2019, lethbridge2004}. Medical refractory CRS is one of the most common indications for outpatient head and neck surgery in adults.  Surgeons treat CRS with functional endoscopic sinus surgery (FESS), a minimally invasive surgery entirely visualized by a rigid endoscope to access the nasal sinuses through the nostrils with slender instruments. The goal of FESS is to restore paranasal sinus aeration and drainage through bone and tissue fragment removal.

Challenges of FESS arise from the narrow field of view of the endoscope and complexity of nasal geometry forming in a semi-stochastic development pattern. This limits visualization of surgical landmarks and proximity to critical structures including eye, brain, and cranial nerves.  As endoscopy alone does not provide sufficient context of patient-specific anatomy, surgeons rely on intraoperative navigation via preoperative imaging (depicting unmodified anatomy) to ensure the achievement of surgical objectives while minimizing the risk of catastrophic patient injuries. Intraoperative navigation provides camera position with reference to the preoperative CT, offering the surgeon real-time understanding of tool-tissue interactions in the surgical scene.

\begin{figure}[h!]
    \centering
    \includegraphics[width=0.8\textwidth]{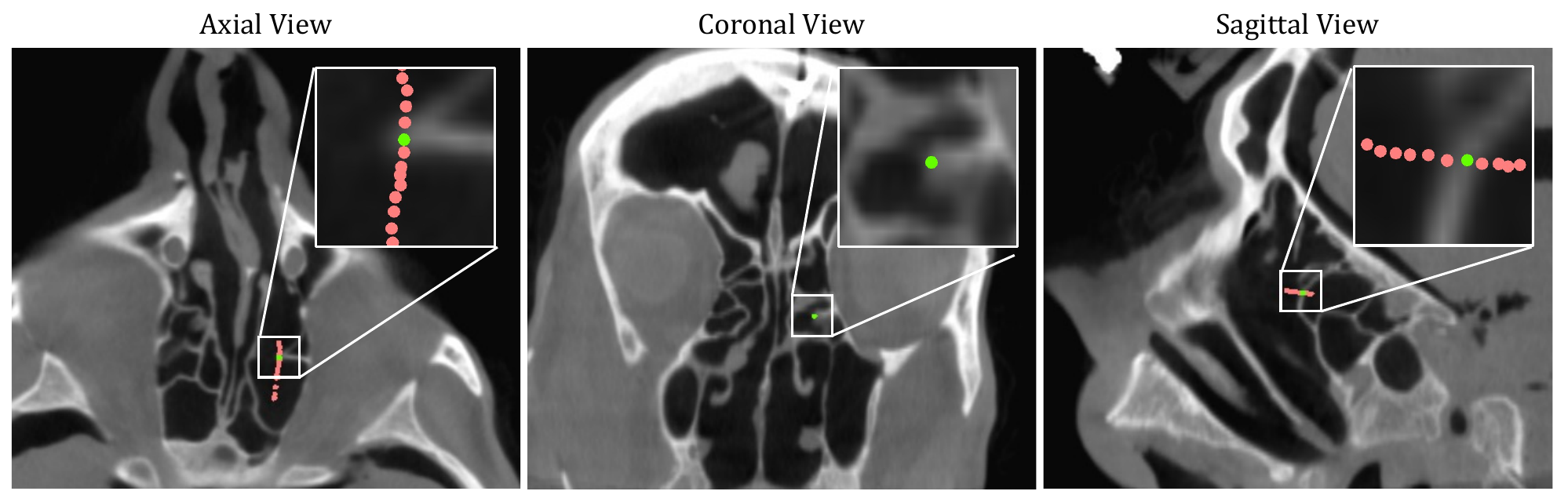}
    \caption{Intraoperative camera positions as recorded via optical tracking in the preoperative CT, shown in tissues that were removed during surgery.  The pink points show the endoscope path through the nasal cavity, while the green point is where the path violates the CT information, as it is in a region where there previously was tissue but is now vacant for the endoscope to pass through.}
    \label{fig:visual_inaccuracy}
\end{figure}

However, the preoperative anatomical representation is limited in its inability to reflect the anatomical changes derived from the surgical intervention. 
FESS is inherently ablative, meaning preoperative imaging loses correspondence with patient anatomy as the surgery progresses. While correspondence can be reestablished with intraoperative imaging, this requires additional radiation exposure, is expensive, and prolongs surgical case times.
This generates a misleading visualization of the anatomy and inaccuracies reflected in intraoperative navigation (Figure \ref{fig:visual_inaccuracy}).   
Hence, there is a need for alternative solutions to visually represent the changes produced during surgical progression.
A natural choice is to rely on the readily available image information provided by the endoscope, as it is a fundamental part of the surgical procedure and brings implicit information about the surgical scene; information that, combined with the optical tracking, can be exploited to update the preoperative model of the anatomy. Our work presents a first vision-based approach to reflect anatomical changes produced by intervention in the anatomical model, which we posit has further implications towards endoscope-centered solutions for improved navigation in sinus surgery.

\noindent \textbf{Related Work:}
Previous works have proposed monocular depth estimation models for endoscopic sequences~\cite{liu2019dense}. Similarly, these sequences have been employed to recover a 3D surface of the sinus anatomy~\cite{liu2020reconstructing, liu2022sage} and in endoscopic camera relocalization~\cite{hernandez2023investigating}, suggesting the feasibility to employ the direct output of the endoscope to recover anatomical structure. Additionally, initial approaches for model updating have been presented in~\cite{chen2021augmented, hu2007intraoperative, miga2000vivo, franccois2021image}. However, these models aim to represent tissue deformation~\cite{chen2021augmented, hu2007intraoperative, miga2000vivo} and do not account for tissue removal, or make fixed depth assumptions~\cite{franccois2021image} that prevent its application to complex anatomical changes like the ones produced by FESS.

Moreover, the concept of intraoperative model update is closely related to the paradigm of digital twins. Digital twin systems have been developed to model real-world processes in a virtual counterpart. In the medical context, digital twins can provide intraoperative guidance by employing computational techniques to capture surgical procedures on the virtual model in real-time~\cite{chalasani2016concurrent, wang2017force,yasin2020evaluation}. This concept has been presented for liver tumor resection~\cite{shi2022synergistic} and skull base surgery~\cite{munawar2023fully, shu2023twin}, but has yet to be explored for FESS. This motivates the development of methods for intraoperative anatomical model generation that account for the changes in sinus interventions.


\noindent \textbf{Contributions:}
Motivated by the recent progress on monocular depth estimation and sinus surface reconstruction, we propose a method to update a preoperative 3D model of the sinus anatomy from intraoperative endoscopic sequences.
Our approach generates a preoperative model employing volumetric fusion of RGB-D images~\cite{zachglobally2007, curlessvolumetric1996, lorensenmarchingcubes1987, zeng20163dmatch}, using depth rendered from preoperative CT volumes and camera poses.
As a result, we build a truncated signed distance function (TSDF) representation of the preoperative CT.
 Our method leverages the information provided by the intraoperative endoscopic video, using sequences depicting the region of interest immediately after surgical modification and tools are removed from the frame. Note that this approach relies on endoscopic frames depicting the sinus cavity without tools. We use the notion of sequences to simulate endoscope input and evaluate the feasibility of a vision-based algorithm for model updating.
We utilize a learning-based monocular depth estimation~\cite{liu2020reconstructing} to project intraoperative depths into the preoperative model, allowing us to estimate the area of change by depth comparison. The estimated intraoperative depths at the modified regions are then substituted in the preoperative TSDF volume to generate an intraoperative 3D structure.  Our implementation further supports sequential updates where we employ the existing TSDF with depth estimates from the next intraoperative step. 




\section{Methods}\label{sec:methods}

 Our method employs a discrete estimation of the truncated signed distance function $D(\mathbf{X})_t$ to represent the anatomy at a given surgical step $t$. 
These surgical steps correspond to the intermediate stages during the resection of the lamella during a simulated FESS procedure. Each step depicts the lamella region of interest after a tissue fragment was removed. The \textbf{Update} process changes the model at step $t$ based on endoscopic video and camera pose information from the next intraoperative surgical step $t+1$. We model this process as:

\begin{equation}\label{eq:model_update}
     D(\mathbf{X})_{t + 1} = \textbf{Update}(D(\mathbf{X})_t, \mathbf{I}_{t+1}, \mathbf{T}_{t+1})
\end{equation}

\begin{figure}[t!]
    \centering
    \includegraphics[width=0.85\textwidth]{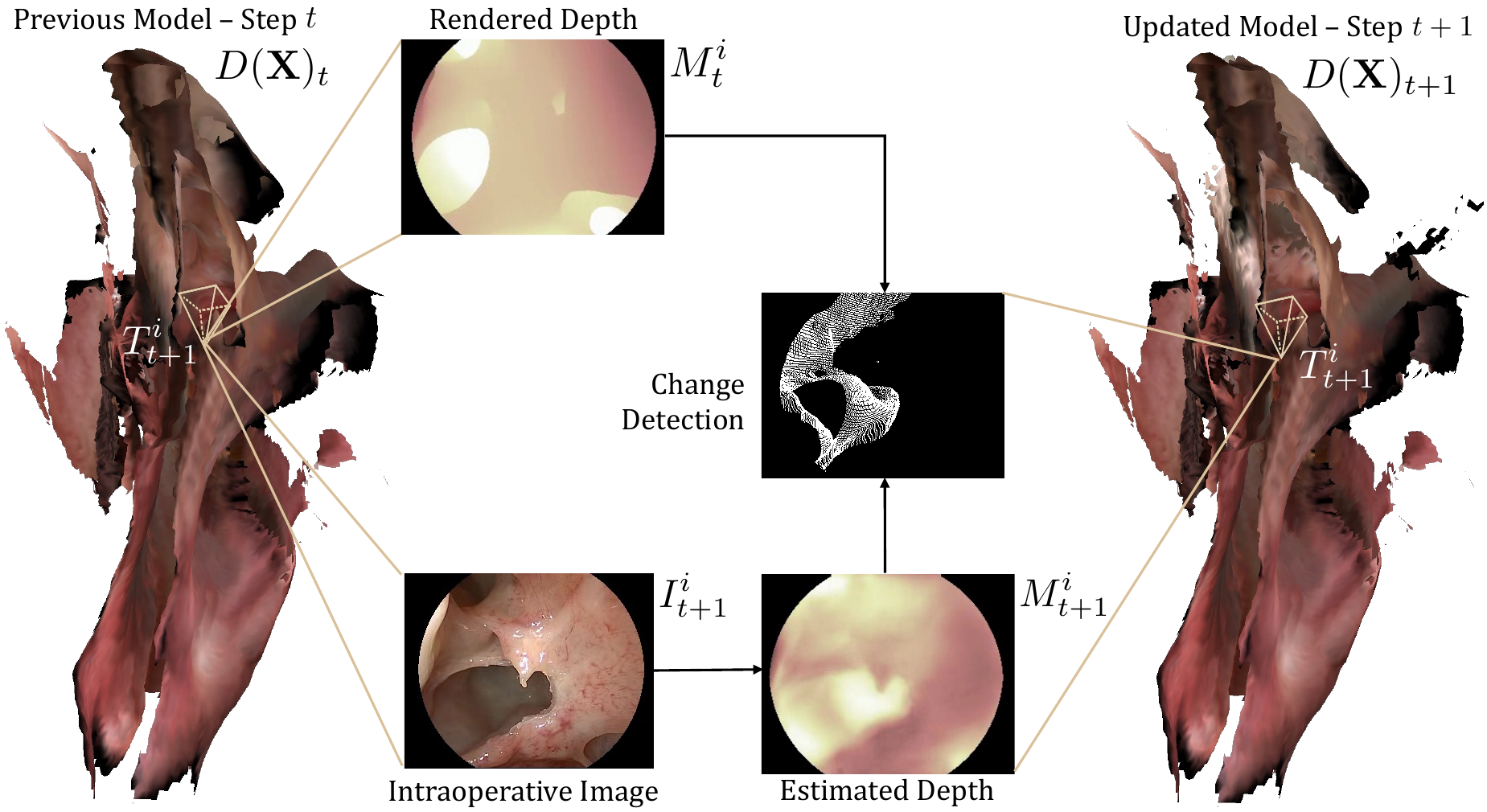}
    \caption{Overview of method for intraoperative update.  Note that the preoperative model is used at surgical step $t = 0$ and depths are rendered from the previous model ($D_(\mathbf{X})_t$) using the camera pose $i$ from the current intraoperative sequence (i.e. $T_{t+1}^i$)}
    \label{fig:overview}
\end{figure}

Where $D(\mathbf{X})_t$ is the previous anatomical model, $\mathbf{I}_{t+1}$ is the current intraoperative endoscopic sequence, and $\mathbf{T}_{t+1}$ camera pose information associated to each frame of $\mathbf{I}_{t+1}$.  The initial preoperative anatomical model $D(\mathbf{X})_0$ is generated in a preoperative step where $t=0$, employing CT, endoscopic video, and pose information. The preoperative data depicts the anatomy prior to any surgical modification. The \textbf{Update} process extracts the region of interest where changes occurred and updates the previous reference model using an update rule inspired by depth fusion methods for truncated signed distance function integration. Note that the endoscope camera was calibrated to extract intrinsic and distortion parameters, and all endoscopic images were undistorted prior to use in our algorithm. An overview of our model update method is shown in Figure{~\ref{fig:overview}} and each step is described in detail in the upcoming sections.


\subsection{Preoperative Anatomical Model}\label{sec:preop_model}
The preoperative model provides a visual reference of the undisturbed anatomy 
and an editable representation of the anatomy that will store the progressive updates.
 We rely on input from a preoperative CT segmentation and an initial endoscopic sequence (RGB images and corresponding camera poses), where the anatomy has not been modified. We use the data collection setup of~{\cite{mangulabnan2023quantitative}} which employs optical tracking of the anatomy and endoscope. We register the tracker to CT space using the segmented anatomy markers visible in the CT, then employ a checkerboard-based hand-eye calibration{~\cite{strobloptimal_handeye2006}} to define the endoscope relative to the CT.

 The preoperative model is generated by employing a depth fusion method{~\cite{curlessvolumetric1996,zachglobally2007}}  to integrate rendered CT depth into a TSDF representation, where the 2D image plane from the camera perspective represents the line of sight of the sensor to the surface. This process requires computing the distances from each camera $i$ to the visible 3D voxels $\mathbf{X}$ in the TSDF, to then extract the distance $d_i(\mathbf{X})$ from the voxel to the anatomical surface based on the depth map (i.e. distance between camera and surface). For further information related to our TSDF representation, refer to Supplementary Figure~\ref{fig:tsdf}. These distances ($d_i(\mathbf{X})$) are obtained based on the camera view to only consider voxels $\mathbf{X}$ that are in the line of sight and properly localize the depth information with respect to each voxel.

 Considering the anatomy segmentation of the CT scan and the endoscope sequence ($\mathbf{I}_{0}$) with camera poses ($\mathbf{T}_{0}$) registered to the CT space, we render depth at every pose $T_0^i \in \mathbf{T}_{0}$ by back projecting every pixel position of the image into the segmented CT surface, generating a depth map $M_0^i$. Note that this process requires the endoscope's camera calibration matrix $K$.   To obtain the distances, we project the 3D voxel into the camera plane employing $\mathbf{X}_c^i = T_0^i \mathbf{X}$, and taking the $z_c$ component of $\mathbf{X}_c^i$ (distance of $\mathbf{X}$ to the camera $i$). Then, we compute the 2D projection $\mathbf{x}$ of $\mathbf{X}$ into the image plane to obtain the desired distance of the 3D voxel $\mathbf{X}$ to the surface in the depth map, as seen from the camera perspective $\mathbf{T}_0^i$. This computation is done by employing $d^{*}_i(\mathbf{X}) = M_0^i(\mathbf{x}) - z_c$. Following~\cite{zachglobally2007, liu2020reconstructing},  we employ the truncated signed distance defined as $d_i(\mathbf{X}) = \min(1, d^{*}_i(\mathbf{X}))$.
Note that $d_i(\mathbf{X})$ represents the distance of the 3D voxels $\mathbf{X}$ to the surface from the perspective of the camera $i$, and not all voxels $\mathbf{X}$ will be visible from each camera pose. For the initialization of the 3D structure, this corresponds to depths rendered from the preoperative CT segmentation. Considering that we do not have intraoperative CTs and therefore cannot rely on depth renders during surgical progression, we employ a learning-based model to generate depth maps from endoscopic images for the subsequent steps. To obtain the TSDF representation, we integrate over all the cameras employing Eq. \ref{eq:depth_fusion}, with $w_i(\mathbf{X}) = 1$~\cite{curlessvolumetric1996,zachglobally2007}.

\begin{equation} \label{eq:depth_fusion}
     D(\mathbf{X})_t = \frac{\sum w_i(\mathbf{X})d_i(\mathbf{X})}{\sum w_i(\mathbf{X})}
\end{equation}
We employ a discretized version of $D(\mathbf{X})_0$ and the cumulative weighting $W(\mathbf{X})_0 = \sum w_i(\mathbf{X})$ as a computational model to store the subsequent intraoperative updates.  This discrete voxel representation allows us to extract the surface corresponding to $D(\mathbf{X})_t = 0$ using Marching Cubes, generating a mesh of the sinus anatomy. For the preoperative model, we recover a colorized 3D surface reconstruction of the anatomy employing  $D(\mathbf{X})_0$ and the endoscopic frames~{\cite{zeng20163dmatch}}.


\subsection{Model Update}
The intraoperative update follows two steps: First, we compare depth information between the surgical steps $t$ and $t+1$ to identify the regions where the tissue changed. Then, we employ a modification of the TSDF integration method to perform the model update. 
We assume that all the intraoperative camera poses are defined with respect to the preoperative CT space corresponding to $t=0$ in the surgical progression, which can be obtained when a navigation system is employed.

\subsubsection{Change Detection}
We identify the regions of the anatomy that require update per frame by comparing the depth information between two consecutive surgical steps. This is motivated by the hypothesis that tissue ablation should also results in modifications in the depth maps of the operated region.  To compare the surgical steps, we first employ the surface reconstruction encoded in $D(\mathbf{X})_t$ to synthesize depth maps $\mathbf{M}_t$ with respect to the camera poses of the next step of the progression $\mathbf{T}_{t+1}$. This yields information about the structure of the anatomy before performing the next anatomical change in step $t + 1$. Given that the 3D model for step $t + 1$ is not available yet, we employ an image-based monocular depth estimator{~\cite{liu2020reconstructing}} that was fine-tuned on the preoperative sequence to generate depth maps for the next step $\mathbf{M}_{t+1}$, as this model has been shown to have state-of-the-art performance for sinus endoscopy. This will generate two sets of depth maps, $\mathbf{M}_t$ and $\mathbf{M}_{t + 1}$, both associated with the same set of camera poses $\mathbf{T}_{t+1}$.


 Then, we employ the information of the depth maps to generate two point clouds $\mathbf{C}_t$ and $\mathbf{C}_{t+1}$ that represent the same anatomical region at different surgical steps, where a surgical step occurs when a fragment of tissue is removed from the anatomy. This is done by processing the depth maps per frame, noting that both depths from the previous and current models share the same camera poses, and hence the same image frames. We solve for the scale due to the ambiguity in monocular depth prediction, and then register the point clouds using the iterative closest point (ICP) registration algorithm with random sample consensus (RANSAC) as implemented by the Open3D Python library{~\cite{zhou2018open3d}}. We use RANSAC considering that the 3D points corresponding to anatomical change are not expected to align between the point clouds, and this method allows us to optimize registration without these outliers. This method also enables us to compensate for soft-tissue deformation.

Once we have registered the point clouds, we recompute the 2D projection of $\mathbf{C}_{t+1}$ into a given camera pose to generate a pair of overlapping and registered depth maps $M_t$ and $M'_{t+1}$. Due to the projections and scaling applied to $M'_{t+1}$, we can lose some depth points and hence we perform the imputation of the missing values with information from $M_{t}$ at the missing location. This process is applied to every depth map of the selected sequence to generate a set of scaled and registered depth maps $\mathbf{M}'_{t+1}$ that lie at the same space as the surface encoded by $\mathbf{D}(\mathbf{X})_t$.  The 2D projection yields estimated intraoperative depth maps aligned at the camera pose that depicts the corresponding anatomy from the previous model. This allows us to compare depths, extract a corresponding binary mask depicting the region of change, and fuse the intraoperative depths at specific regions, relative to the camera.

\begin{figure}[h!]
    \centering
    \includegraphics[width=0.85\textwidth]{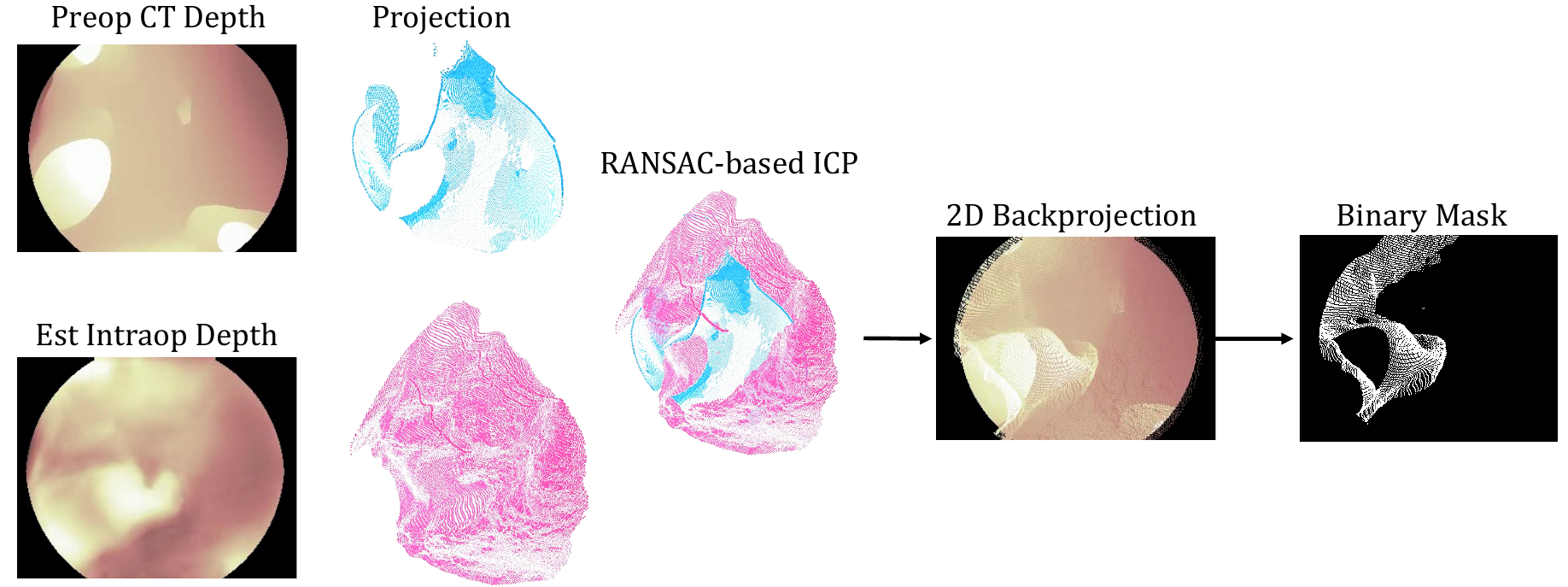}
    \caption{Framewise change detection.}
    \label{fig:change_detection}
\end{figure}

The scaled and registered depth maps $\mathbf{M}'_{t+1}$ are compared with its counter parts in $\mathbf{M}_{t}$ to compute the pixelwise difference between the surgical steps. Finally, these differences are thresholded to extract a binary mask $\mathbf{m}_{t+1}$ pertaining to the region of interest for each frame in $\mathbf{I}_{t+1}$.  We assume that the changed regions will have a larger depth than the depth render of the previous model since tissue is always removed in each surgical step of our simulated procedure. Thus, we set the threshold based on estimated depths larger than $1.0$~mm from the pre-existing surface to discount minimal errors from the registration. This process is depicted in Figure \ref{fig:change_detection}. 

\subsubsection{Intraoperative Model Generation} \label{sec:ct_update}

The change detection step outputs a binary mask indicating the regions of $D(\mathbf{X})_t$ that requires updates. To generate the updated TSDF with the new changes, we redefine the weighting factor of the distance integration of Eq. \ref{eq:depth_fusion} to consider the changes indicated by the binary change mask. 
We thus define the update rule in terms of the incremental calculation in~\cite{curlessvolumetric1996}. Considering a set of truncated signed distances $d_i(\mathbf{X)}$ calculated employing the poses $\mathbf{T}_{i+1}$ and registered depth maps $\mathbf{M}_{t+1}$, employing a process similar to 
section \ref{sec:preop_model}, the updated anatomy representation $D_{t+1}$ is computed incrementally with the integration rule in Eq. \ref{eq:fusion_update} 
(the $\mathbf{X}$ is omitted for simplicity).
\begin{equation} \label{eq:fusion_update}
    D_{t+1, i+1} = \frac{W_{t+1, i}D_{t+1, i} + w_{t+1, i+1} d_{i}}{ W_{t+1, i} + w_{t+1, i+1} }
\end{equation}
with $W_{t+1, i+1} = W_{t+1, i+1} + w_{t+1, i+1}$. The weight and the incremental distances are initialized with the parameters of the previous model as $W_{t+1, 0} = W_t$ and $D_{t+1, 0} = D_t$, and all $w_i = 0*W_{t} + \mathbf{m}_{i,t+1}$ are defined in terms of the previous weights and the current change mask. The process iterates over all the depths in $\mathbf{M}'_{t+1}$. The binary mask in the weighting parameter reduces the contribution of the previous representation in favor of the new depths for the cases where $\mathbf{m}_{i,t+1} = 1$ (regions of anatomical change), while keep distances close to their initial values in the cases the tissue was not perturbed. 


\section{Experiments and Results}\label{sec:experiments_results}
We evaluate the results of our method through five intraoperative perforations (or bites). These bites correspond to the first tissue ablation of a simulated FESS performed by an experienced surgeon on an ex vivo specimen. For each step, we capture endoscopic video, optical tracking, and CT data employing a setup composed of the Storz Image1 HD camera (Karl Storz SE \& Co. KG, Tuttlingen, Germany), NDI Polaris Hybrid Position Sensor (Northern Digital Inc., Waterloo, Canada), and Brainlab LoopX scanner (Brainlab, Munich, Germany)~\cite{mangulabnan2023quantitative}.  We use input intraoperative endoscopic sequences that contain 70-90 frames based on the longest continuous sequence that depicts the region of interest to ensure adequate information is employed for the update. We performed an additional experiment to further support this choice, which verified that increasing sequence length reduces error between the updated mesh and ground-truth CT. Additional details are reported in Supplementary Figure~\ref{fig:seq_length}.

\subsection{CT Comparison}
\begin{figure}[t!]
    \centering
    \includegraphics[width=0.75\textwidth]{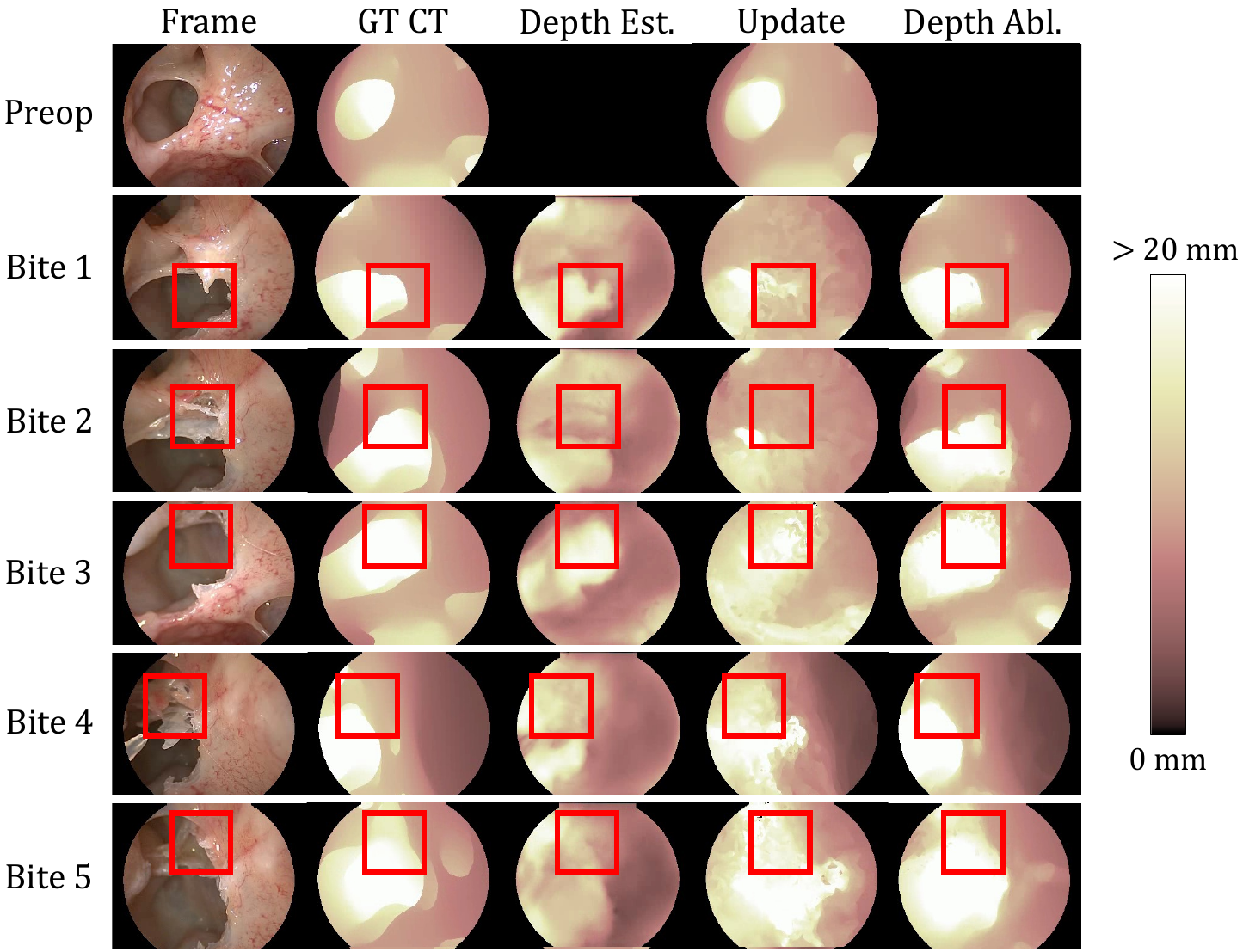}
    \caption{Endoscopic video frames depicting anatomical change with corresponding depths from preoperative and intraoperative CTs, estimation, and rendered from the update and depth ablation meshes.  The regions modified between each intraoperative step are highlighted in red. Note that the depths are rendered from the associated camera pose of the frame, and these poses differ between sequences.}
    \label{fig:bites_results}
\end{figure}

The generated intraoperative meshes were evaluated with respect to the ground-truth CT segmentation after each surgical perforation. The anatomical changes are visualized in Figure \ref{fig:bites_results}.

Considering that the updated mesh is rendered with respect to the preoperative CT coordinate frame, each intraoperative ground truth CT was registered to the preoperative CT for comparison.  The intraoperative CT was initially transformed by performing a 3D rigid registration between segmented anatomy markers in both CTs. The registration was then refined using an ICP algorithm using depths rendered at camera poses depicting no anatomical changes, where points were sampled across this sequence, resulting in residual errors of 1.832 mm.
A 3D bounding box pertaining to the anatomical change was then defined based on the maximum amount of change in the last surgical step to compute errors within a constant region of the anatomy across surgical progression.
Depth was then rendered at a set of camera poses viewing the region of anatomical change from the preoperative CT, registered intraoperative CT, and updated model.  These camera poses were kept constant for each intraoperative step to localize the evaluation, as unchanged regions of the mesh may bias the comparison since more of the anatomy is modified during surgical progression. Furthermore, we rely on the camera poses rather than a direct comparison of the mesh to establish correspondences, considering that the closest points between the meshes may not necessarily represent the same part of the intricate sinus anatomy. The resulting depth maps were then projected to 3D to compute the error between point correspondences within the 3D bounding box, confining our evaluation to a constant set of points between sequences. The mean errors of the preoperative CT where no update was employed and updated meshes with respect to the intraoperative CT are shown in Table \ref{tab:results}. The mean difference between the pre- and intraoperative CT represent the upper bound of errors, as the anatomical changes are not represented preoperatively.

\begin{table}[t!]
        \centering
        \caption{Mean error between reference GT CT and different 3D anatomical models.}
        \label{tab:results}
        \begin{tabular}{c|cccc}
     &  \multicolumn{4}{c}{Mean $\pm$ Std Error (mm)}\\
            \hline
             Reference&No Update&Updated& Depth Ablation&Registration Ablation\\
             Ground Truth CT&Preoperative CT&Model&Model&Model\\
             \hline
             Preoperative&  -& $0.233 \pm 1.3$& -&-\\
             Bite 1&  $1.875 \pm 3.7$& {$2.432 \pm 2.3$}& $1.795 \pm 2.2$& {$6.317\pm 3.2$}\\
             Bite 2&  $4.095 \pm 5.0$& {$3.397 \pm 3.7$}& $1.804 \pm 3.0$& {$5.455\pm 2.5$}\\
             Bite 3&  $4.573 \pm 5.0$& {$3.172 \pm 3.6$}& $2.250 \pm 2.6$& {$5.280\pm 3.2$}\\
             Bite 4&  $4.634 \pm 4.8$& {$3.538\pm 2.8$}& $1.778 \pm 1.6$& {$6.926\pm 4.0$}\\
             Bite 5&  $5.567 \pm 4.6$& {$3.280 \pm 2.6$}& $2.717 \pm 1.7$& {$9.580\pm 5.1$}\\
     \end{tabular}
    \end{table}



\subsection{Ablations}
\bmhead{Depth Estimation}
We ablate the depth estimator to evaluate the influence of its accuracy. The depths were rendered from the ground-truth intraoperative CT to simulate reliable depth estimates and were then registered to the preoperative model to detect anatomical modification. The intraoperative CT depths were then used as input to the TSDF at the regions of change. The resulting depth differences from the generated mesh are shown in Table \ref{tab:results} - Depth Ablation Model and a visual of rendered depths in the last column of Figure \ref{fig:bites_results}.

\bmhead{ICP Registration}
The estimated depths are first registered to the preoperative model using a RANSAC-based ICP algorithm prior to re-integration of the TSDF volume.  This registration intends to account for potential error in the alignment of the anatomy with respect to the camera pose, as misalignment in anatomical boundaries can present large depth differences that would significantly affect the localization of the change. We ablate the use of this component to further assess the value of this step in generating the intraoperative mesh. Given the inherent differences of monocular depth estimation, the frames were normalized to the preoperative depth render to detect change and update the TSDF representation. The results of this ablation are shown in Table \ref{tab:results} - Registration Ablation Model.

\section{Discussion}\label{sec:discussion}
We compare the difference between the intraoperative CT when no update was employed (preoperative CT) and with the updated model using our method based on distance errors in the region of anatomical change. As the preoperative CT was used to generate the initial model, we expect small error from this mesh (preoperative row in Table \ref{tab:results}) although some discrepancies may be introduced during 
surface extraction, for example, due to the endoscope's limited field of view which might not provide enough depth information. We expect the error of the preoperative CT to increase with respect to the intraoperative CT during progression, providing an upper bound to the errors 
since anatomical changes will cause greater differences in depth. The first intraoperative bite increases in error, while the remaining steps decrease. This increase may be due to the quality of the depth estimation as well as the intraoperative CT segmentation. The registration between the preoperative and intraoperative CT used for evaluation may contribute to the error, however, these are likely to be minimal considering the residual errors on unmodified regions of the anatomy.

We ablate the depth estimator using depths rendered from the intraoperative CT to update the model with reliable values
(Table \ref{tab:results} - Depth Ablation Model). These models are shown to have minimal errors meaning that the update is limited by the accuracy of the depth estimation. The small margin of error in the depth ablations may be a result of the variation between the segmentation of the CT scan, further limited by the resolution of the CT scanner.
The segmentations may also vary in thickness, meaning that the boundaries of the tissue walls may fluctuate as these were produced independently of each other due to partial volume effects in the CT.

Considering that each intraoperative frame is transformed to the preoperative TSDF prior to updating the representation, misalignment in the CT segmentations may contribute to the update error as the mesh is aligned in the preoperative space. 
We ablate the use of registration in the model update to evaluate the effectiveness of this step
(Table \ref{tab:results} - Registration Ablation Model). We find that the updated meshes without this step present higher errors, meaning that registering the depth estimates to the preoperative model is a key step for the process

Our method depends on known endoscopic camera poses to render depth and localize the intraoperative frames in CT space. 
This simulates current navigation systems used in FESS, where the preoperative CT is given, and the model update could potentially be improved to leverage more information from CT scan to optimize specifically for this procedure. The purpose of FESS is to remove tissue, meaning that depth could potentially be recovered when utilizing the entire CT in the initial TSDF volume. As surfaces are exposed intraoperatively, the modified depths would already be represented and we could hypothetically check for occupancy and the detected changes to remove values associated with ablated tissue. 

Our approach generates the initial mesh based on an initial scoping to render depth from the preoperative CT. While this may be limited by the range of the corresponding endoscopic sequence, this method lends itself to employ alternative sources to the preoperative model rather than the CT scan, such as vision-based reconstruction approaches from Structure-from-Motion (SfM) or Simultaneous Localization and Mapping (SLAM) algorithms.~\cite{liu2020reconstructing} and~\cite{liu2022sage} have shown that it is possible to generate sinus reconstructions directly from endoscopic video, where these structures are also limited by the range of the scoping meaning that an entirely vision-based solution for FESS navigation would maintain this limitation. Considering that SfM and SLAM methods also produce camera poses of each frame with respect to the structure, our method could generate the initial preoperative TSDF representation with these reconstructions. Camera relocalization would also be necessary to define the intraoperative camera poses relative to the reconstruction for model updating, however,~\cite{hernandez2023investigating} has shown that this problem requires further investigation to produce reliable pose estimations in the sinus. This motivates future work in camera relocalization to enable the substitution of a vision-based reconstruction in this method.


\section{Conclusion}\label{sec:conclusion} 
This work presents a method for updating a preoperative 3D anatomic model for navigated sinus surgery using intraoperative endoscopic video. 
Our results indicate that this approach is feasible 
to reduce errors during surgical progression contrary to an increase 
when not updating. Future work will be focused on refining the monocular depth estimation considering the improvement in the updated model with ground-truth depth, and eliminating the dependency on external navigation systems towards a fully endoscope-centered solution. This first vision-based approach to preoperative model updating not only enhances surgical planning and intraoperative navigation, but also paves the way for the realization of a digital twin for sinus endoscopy.

\backmatter

\bibliography{references}

\section*{Supplementary Material}

\subsection*{Truncated Signed Distance Function}
\begin{figure}[h!]
    \centering
    \includegraphics[width=0.6\textwidth]{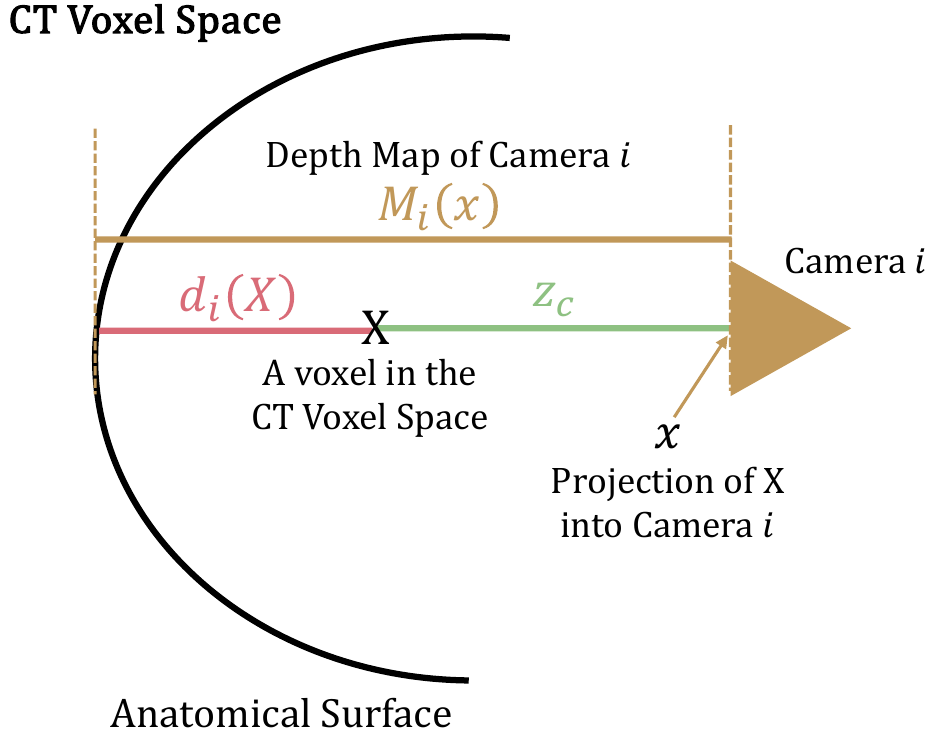}
    \caption{Truncated signed distance function representation using a discrete voxel grid of the CT. We employ the cameras and corresponding depth maps to extract the distance from each voxel to the anatomical surface.}
    \label{fig:tsdf}
\end{figure}

We employ a truncated signed distance function (TSDF) based off of{~\cite{curlessvolumetric1996}} to serve as an editable representation of the sinus anatomy for intraoperative updates. This discrete voxel grid implementation of TSDF enables us to define each signed distance field as the distance of each voxel in the CT space $\textbf{X}$ to the surface point in the line of sight to the endoscope camera $i$. We can then extract the distance $d_i(\textbf{X})$ by projecting the voxel $X$ into camera $i$ and employing the corresponding depth map $M_i(x)$. A visual representation of this distance is shown in Figure{~\ref{fig:tsdf}}.

\subsection*{Effect of Sequence Length}

\begin{figure}[h!]
    \centering
    \includegraphics[width=0.9\textwidth]{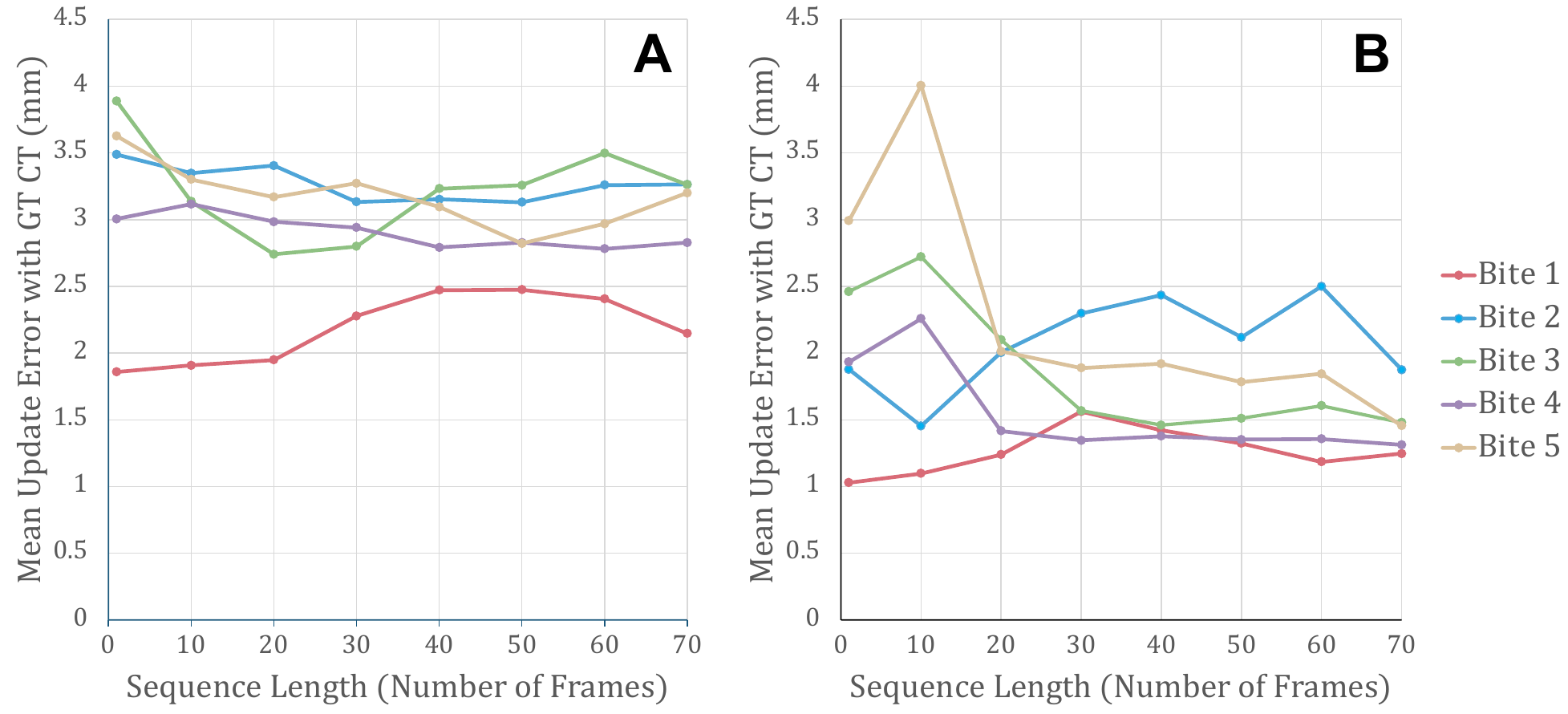}
    \caption{Mean update error with the ground-truth CT for variable sequence lengths based on distance between point correspondences (mm) where (\textbf{A}) compares the intraoperative meshes generated with estimated depth maps and (\textbf{B}) with ground-truth depth renders from the intraoperative CT.}
    \label{fig:seq_length}
\end{figure}

We evaluated our proposed update algorithm by using intraoperative sequences, simulating input from the endoscope camera during surgery. These sequences were determined by the longest continuous sequence of frames depicting the anatomical region that was modified, where there are no tools present in the frame. We report that longer sequences generally reduce the error between the updated mesh produced by our algorithm and the ground-truth CT as shown in Figure~{\ref{fig:seq_length}}.

The results shown in Figure~{\ref{fig:seq_length}} (\textbf{A}) correspond to the error where we use estimated depth maps from a learning-based monocular depth estimation for the update. Considering that each frame may have variable accuracy and may cause these errors to fluctuate depending on which part of the sequence is sampled, we also present the change in error for the depth ablation experiment where the depth maps are rendered from the ground-truth CT shown in Figure~{\ref{fig:seq_length}} (\textbf{B}). This experiment allows us to verify that longer sequences generally improve the fidelity of the update, as more intraoperative information can be employed in our algorithm.

\end{document}